\documentclass[12pt]{iopart}
\usepackage{graphicx}

\usepackage[backend=bibtex,style=ieee,natbib=true]{biblatex} %added
\usepackage[colorinlistoftodos,prependcaption,textsize=scriptsize]{todonotes}

\newcommand{\ideas}[2]{}{}

\begin{document}

\title{From Natural to Artificial Camouflage: Components and Systems}

%%%%% Alternate titles
%
% Sensing, Actuation and Distributed Computation in Natural and Artificial Camouflage
%
% From Natural to Artificial Camouflage: Components, Algorithms, and Systems

\author{Yang Li and Nikolaus Correll}

\address{Department of Computer Science, University of Colorado Boulder, Boulder, USA.}
\ead{\{yang.li-4, nikolaus.correll\}@colorado.edu}
%\vspace{10pt}
%\begin{indented}
%\item[] October
%\end{indented}

\begin{abstract}
	% Leave until Nikolaus review the whole paper.
We identify the components of bio-inspired artificial camouflage systems including actuation, sensing, and distributed computation. After summarizing recent results in understanding the physiology and system-level performance of a variety of biological systems, we describe computational algorithms that can generate similar patterns and have the potential for distributed implementation. We find that the existing body of work predominately treats component technology in an isolated manner that precludes a material-like implementation that is scale-free and robust. We conclude with open research challenges towards the realization of integrated camouflage solutions.
\end{abstract}

% Uncomment for PACS numbers
%\pacs{00.00, 20.00, 42.10}
%
% Uncomment for keywords
%\vspace{2pc}
%\noindent{\it Keywords}: XXXXXX, YYYYYYYY, ZZZZZZZZZ
%
% Uncomment for Submitted to journal title message
%\submitto{\JPA}
%
% Uncomment if a separate title page is required
%\maketitle
%
% For two-column output uncomment the next line and choose [10pt] rather than [12pt] in the \documentclass declaration
% \ioptwocol
%

\section{Introduction}
\label{sec:introduction}

% 1. Cite some B&B papers on bio system designs as the bigger question behind our question?
% 2. Overview on integration (cite the Science paper)

Camouflage is a survival skill that animal uses to deceive other animals to hide, and has been extensively researched \cite{stevens2009animal, stevens2011animal, bradbury2011principles}. Animals camouflage themselves in a wide variety of ways using patterns, textures  \cite{pikul2017stretchable}, or shapes  \cite{allen2014comparative} that are suited to their environment. 
\begin{figure*}[!htb]
\centering
\includegraphics[width=.45\textwidth]{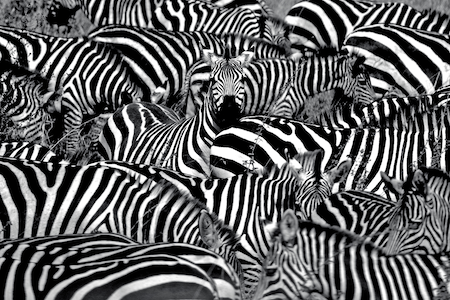}
\includegraphics[width=.45\textwidth]{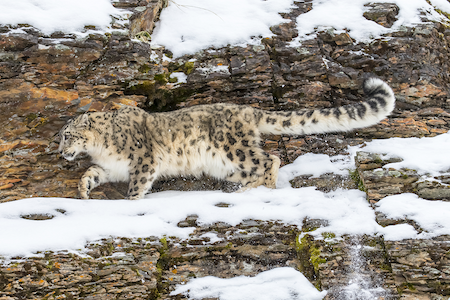}
\\
\includegraphics[width=.45\textwidth]{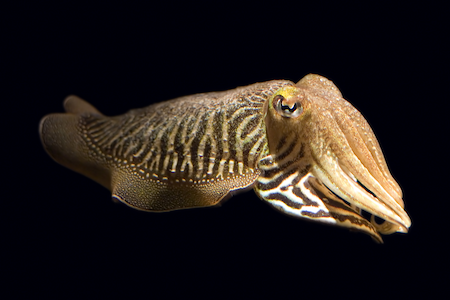}
\includegraphics[width=.45\textwidth]{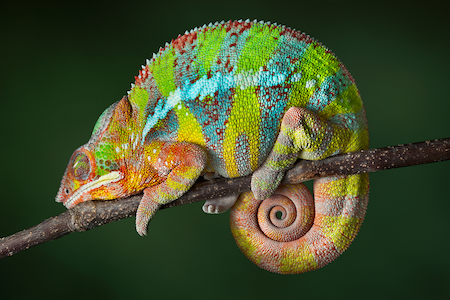}

\caption{Static camouflage examples and dynamic camouflage examples.  \copyright ~Dreamstime.com under license RF-LL. Images of zebra, leopard, cuttlefish, and chameleon are courtesy of Tallllly, Walter Arce, Bill Kennedy, and Cathy Keifer, respectively. }
\label{fig:camouflage_patterns}
\end{figure*}
%\cite{hanlon1988adaptive, stuart2009camouflage}
%Animal coloration serves multiple purposes, including camouflage, signaling (especially mate choice), and thermo-regulation~\cite{stevens2016color}.
Some animals maintain a static coloration pattern during their whole life --- for example, zebras and leopards (see Figure~\ref{fig:camouflage_patterns}, A and B). Others can dynamically change their appearances in different environments, like cephalopods and chameleons (see Figure~\ref{fig:camouflage_patterns}, C and D). Coloration is created by pigment cells, known as chromatophores in fishes, amphibians, and reptiles, and melanocytes in mammals and birds. Static patterns are formed during an animal's development where pigment cells %with constant color
migrate toward a specific position on the body. Several kinds of animals will evolve their skin patterns during their growth from juvenile to adult, which has been acutely documented for lizards~\cite{manukyan2017living}. If ablated, pigment cells can also regenerate, although not necessarily recreating the exact same pattern, but a similar one providing cues for the underlying mechanism's workings. For example, the skin of zebrafish can regenerate new chromatophores after parts of skin were ablated~\cite{yamaguchi2007pattern, nakamasu2009interactions}, closely matching the original pattern, albeit with chromatophores at different locations. Dynamic patterns, instead, are formed using pigment cells that can selectively display specific colors --- for example, a cephalopod's chromatophore changes color by an annular muscle that opens and closes, thereby revealing underlying pigment.

As these muscles require control by a nervous system, the community distinguishes growth-based, or morphological, and nervous-system based, or physiological pattern formation ~\cite{fujii2000regulation, stuart2009camouflage}. Morphological color change is slow and long lasting (days, months, or even years), and is believed to be under hormonal control and inter-cellular interactions. Physiological color change can happen in the order of seconds and is believed to be controlled by the nervous system~\cite{messenger2001cephalopod, nery1997pigment}. Stevens states in~\cite{stevens2016color} that ``Color change can involve modifications that occur through physiological color change, often thought to involve the contraction and dispersion of colored pigment within chromatophore cells, longer-term changes relating to morphology in cellular distribution and pigment synthesis, and in development.''

The processes that lead to camouflage with physiological color change or physiological camouflage, are complex and involve information processing throughout the body. This process starts with perception in the eye, decision making in the brain, signaling through the nervous system, and ends with pattern formation on the skin. While the exact processes are not fully understand, it is important to note that the skin is not simply a display, but limited to generate specific classes of patterns suggesting low-bandwidth encoding of information traveling from the brain to the skin. Furthermore, there is evidence that cephalopod skins are sensitive to light~\cite{mathger2010evidence, kingston2015visual, ramirez2015eye}, suggesting that perception happens not only in the eyes and the central nervous system, and that there exist light-activated local color change independent of the brain.

This survey is driven by the question how we can engineer artificial camouflage skins, suggesting a system-level view of the natural system and dissecting it into its sensing, actuation, computation, and communication components~\cite{mcevoy2015materials,yang2018grand}. %In the interest of engineering dynamical camouflage systems, we are mostly interested in physiological camouflage. Yet, due to the similarity of the resulting patterns, we are also interested in the underlying mathematical models, which have been mostly studied for morphological pattern formation.
Whereas sensing and actuation have the same meaning in physiology and engineering, computation and communication are the engineering equivalent of information processing in biological systems. In engineered systems, computation, i.e., performing operations on information, is distinct from communication, i.e., moving information throughout the body. We believe that this distinction, and in particular the information theoretical tools that exist for computation and communication, to also help with describing physiological processes in animals, and will do so when appropriate. 

To the best of our knowledge, this is the first survey that looks at the problem from a comprehensive perspective, spanning recent results from biology to materials and computer science. We review mechanisms of animal camouflage, mathematical models describing pattern formation, materials for artificial camouflage, and then discuss possible designs for artificial camouflage systems. We also argue that a constructive approach to engineer artificial camouflage systems might also generate possible hypothesis on the function of biological systems in the spirit of ~\cite{webb2001can, kreit2013biological}.

The remainder of this paper is organized as follows. Section~\ref{sec:mechanism} reviews research on mechanisms of animal camouflage. Reaction-diffusion and cellular automata models for pigment pattern formation are reported in section~\ref{sec:model}. We then report research that shows evidence that these models are indeed sufficient to generate the patterns observed in biological systems. In section~\ref{sec:material}, we describe various artificial mechanisms to create color change and attempts at system-level implementations. We discuss the requirements for designing a proper camouflage system and conclude the survey in section~\ref{sec:conclusion}.

%
%All the above research provides us inspiration and guidance on designing an artificial camouflage system, especially that can perform adaptive camouflage. The design requires the integration of the visual perception, decision making on the pattern, and pattern formation with tens or hundreds of artificial chromatophores. This topic is interdisciplinary and a combination of material engineering, electrical engineering, computer science and mechanical engineering. First, we need to design artificial chromatophores with the abilities to change color or move, and maybe perform sensing like cephalopods' skin. Then, a control mechanism is required to obtain environmental information from other sensors, make decisions, and send control signals to actuate artificial chromatophores.

\section{Mechanism of Animal Camouflage}
\label{sec:mechanism}

In this section, we review the mechanism of camouflage starting from the animal's perception of its environment, to selecting an appropriate pattern, and finally its implementation using chromatophores on the skin.

\subsection{Visual Perception}
\label{sec:visualperception}

Animals that perform rapid adaptive camouflage with physiological color change use visual perception to identify their environment and display appropriate camouflage patterns. Albeit how animals perceive their environment is not fully clear, the consensus is that the majority of the perceived information is visual and processed in the brain~\cite{troscianko2009camouflage}. Visual perception has been extensively studied in the cuttlefish, which is known for its rapid adaptation capabilities~\cite{kelman2008review, chiao2015review}. The visual strategy that cuttlefish use to select camouflage patterns and coloration is not limited to patterns, but includes object recognition to contextualize an environment~\cite{kelman2008review}. The cuttlefish actively processes features like edges, contrast, and size, to identify specific objects and consequently decides upon a camouflage pattern~\cite{stevens2006disruptive, zylinski2009perception, chiao2015review}. In~\cite{zylinski2009perception}, it is shown that discrete objects are important in the cuttlefish's choices of camouflage, and the underlying mechanism is similar to that in vertebrates. Identification of an object, so-called figure-ground processing, has two stages. First, in a low-level process individual neurons detect the locations, polarity, and orientation of small edge segments~\cite{hanlon2009cephalopod}. In a second stage, grouping local edges helps to identify individual objects and background information~\cite{lamme1995neurophysiology, grossberg1997visual}.

\subsection{Patterns for Camouflage}
\label{sec:patternsforcamouflage}

In order to understand what the best possible camouflage pattern is for a specific situation and when to select it, it helps to consider the evolutionary origins of the mechanism. Minimizing the likelihood of detection is one of the major components for adaptive camouflage. In general, edge detection and edge grouping might be exploited in two ways by a prey animal to become less visible to predators. First, if its coloration and brightness is similar to that of the background, also known as background matching, a predator might have difficulties to discern the contours of the animal. A second way of exploiting the mechanisms of edge processing is to disrupt the grouping of the small edge segments to form a coherent outline of a whole object~\cite{troscianko2009camouflage}, which is called disruptive coloration. For example, cephalopods seem to rely on a few simple families of patterns of speckles and stripes that are classified into uniform, mottled and disruptive~\cite{hanlon2007cephalopod}. Uniform patterns exhibit little to no contrast, mottle patterns exhibit small-to-medium scale light and dark patches, whereas disruptive patterns have large-scale light and dark components of multiple shapes, orientations, scales, and contrasts.

There is debate among researchers as to whether background matching, disruptive coloration, or the combination of the two have a more important role in camouflage~\cite{stuart2009camouflage}. This discussion gets further complicated as the efficiency of a camouflage pattern also depends on the perception apparatus of the predator~\cite{stuart2008predator}. Some animals change their appearance in the presence of different predators. Chameleons, for example, tend to exhibit stronger background color matching in response to birds than in response to snakes because snakes have poor color discrimination~\cite{stuart2008predator}. This kind of selective camouflage is also hinting at camouflage being physiologically costly~\cite{stuart2008predator}. This is particularly important for animals capable of rapid color change, for example to perform background matching during motion.

In any case, it is worth noting that a camouflage system does not simply reproduce the environment around it like a display. This might have the opposite effect by standing out even more than a static camouflage pattern. We also note that there is no ``perfect'' camouflage pattern for a certain environment, but depends on the desired goal, the capabilities of other animals, and energy efficiency.

\subsection{Pattern Formation}
\label{sec:patternformation}

In this section we first introduce how patterns are formed during development and growth, and then report proposed mechanisms for dynamic pattern formation.

Most of the animals have their skin patterns formed during development. Controlling the distribution of cellular phenotype through regulation of gene interactions and cell behaviors is achieved by cell differentiation, a process only partially understood. Zebrafish is a common model animal to investigate mechanisms of pigment pattern formation. The zebrafish pattern is generated by combination of different types of chromatophores that are formed from the neural crest and must migrate to reach their final destinations~\cite{kelsh2009stripes}. Identification of the signaling process that guides the migration will be crucial in illuminating the mechanisms of chromatophore pattern formation~\cite{kelsh2009stripes}. Variation in migration, population size, organization, and differentiation of chromatophores within the outer skin generates the diversity of pigment patterns~\cite{kelsh2009stripes}. In~\cite{jernvall2003mechanisms}, the authors investigate the developmental mechanisms in metazoan organisms. During development, cells are constantly sending and receiving molecular signals. The network of transcription factors and transduction molecules within a cell integrates the cell's previous history with received signals and then alters cell behaviors. By modifying selected genes, researchers have also found particular genes that are related to the control of pattern formation --- for example, the zebrafish connexin41.8 gene (Cx41.8) is responsible for the leopard phenotypes~\cite{watanabe2006spot}.

It is assumed that pigment cells exchange information with each other and update their states based on the integrated states of neighbor cells. This process is called morphological pattern formation. Depending on the animal system, this communication process can last for days, months, or even years and varies in frequency.  Several mechanisms for morphological pattern formation are proposed. In~\cite{kelsh2004genetics}, the author proposed that pigment cell patterning might result from long-range patterning mechanism, from local environmental cues, or from interactions between neighboring pigment cells. Jernvall et al. \cite{jernvall2003mechanisms} propose that developmental mechanisms of pattern formation in metazoan organisms can be classified into three categories: cell autonomous, inductive, and morphogenetic. A morphogen is a long-range signaling molecule that acts over a few to several dozen cell diameters to induce concentration-dependent cellular response~\cite{rogers2011morphogen}. The graded morphogen distribution thereby subdivides tissues into distinct cell types that are arranged as a function of their distance from the source~\cite{frankham2010introduction}.

Physiological pattern formation, which is implemented with physiological color change, is fast and neurally controlled, as mentioned in section~\ref{sec:introduction}. It is commonly assumed that the skin is only able to generate a limited set of patterns, which are formed by stimulating different parts of brain neurons~\cite{messenger2001cephalopod}. Instead of controlling each pigment cell individually, there are local networks of neurons that control pattern formation. Here, a single neuron is connected to many pigment cells, which in turn receive input from many neurons. This kind of hierarchical control is also observed in the octopus' motor control, which is highly decentralized~\cite{zullo2009nonsomatotopic,wells2013octopus}.

%\emph{Albeit similar in appearance, it is not fully clear how physiological pattern formation models reaction-diffusion processes in morphological pattern formation. [Do we need this?]}

%\subsection{Chromatophores of Cephalopods}
\subsection{Active color change using chromatophores}
\label{sec:chromatophore}

%The basic component for pattern formation are pigment cells in animals skin. As we mentionsed above that there are mainly two kinds of pigment cells --- chromatophores and melanocytes. 
Chromatophores are pigment-containing and light-reflecting cells, or groups of cells, that are found in a wide range of animals including fish, amphibians, reptiles, and cephalopods~\cite{bagnara1968dermal, florey1969ultrastructure, taylor1970chromatophores}. They are largely responsible for displaying skin and eye color in animals and are generated in the neural crest during embryonic development which makes chromatophores special pigment cells~\cite{schartl2016vertebrate}. Mature chromatophores are grouped into subclasses based on their color under white light and the colors are different between the chromatophores of different species --- for example, fish has up to six types, while mammals have only one pigment cell type, the melanocyte (black, brown, red or, yellow)~\cite{messenger2001cephalopod, kelsh2004genetics}.

In this section, we review the research of color change control and mechanism. Since cephalopods are better studied and recorded than other dynamic camouflage animals, we focus on using cephalopods as the research object. In addition to understanding the underlying physiology, we wish to gain insights into whether an engineered camouflage system should be organized in a centralized or distributed way, that is where information processing should optimally happen.

%We report how the chromatophores are connected and controlled for two reasons. One is to better understand the mechanism underlying the pattern formation and enlighten and examine the proposed models for pattern formation. The other is for the design of artificial camouflage system, in the case that how should the artificial color-changing cells be connected and controlled, centralized or distributed, directly or hierarchically? Definitely, adaptive camouflage needs neural control to quickly adapt skin color to a different environment, on which cephalopods are masters.

Chromatophores of cephalopods are under control of the central nervous system, making them fundamentally different from those of fish, amphibian or reptiles~\cite{florey1966nervous, florey1969ultrastructure, young1974central, messenger2001cephalopod}. Each chromatophore organ consists of a pigmented cell and several radial muscles, whose expansion and retraction leads to different amount of pigment displayed. The chromatophores of cephalopods are controlled by a set of lobes organized hierarchically (from optic lobes to peduncle lobes, to lateral basal lobes, and then to chromatophore lobes) so that cephalopods can adapt their appearance extremely rapidly (within milliseconds or seconds). Conceptually, information flows as follows: visual input from the eyes causes the selection of an appropriate pattern; the decision signals pass through the set of lobes to each chromatophores' motor neurons. Chromatophores expand or retract based on motor neurons' activity or inactivity, displaying the body pattern selected. The chromatophores, however, are not innervated uniformly; specific nerve fibers innervate groups of chromatophores, while each radial muscle is innervated by more than one nerve branch~\cite{messenger2001cephalopod}. Also, differently colored chromatophores are independently innervated~\cite{florey1969ultrastructure}. Although the connections from the brain to each chromatophore are well studied, far less is known about the central control of chromatophores, mainly because of the difficulties of recording neural signals. We do know, however, that cephalopods choose from and recombine a few basic patterns which are ``hard-wired'' into the central nervous system and are not learned~\cite{messenger2001cephalopod}. This suggests that pattern formation itself is a decentralized process, which might be triggered by high-level information from the brain. 

In addition, the skin of cephalopods contains molecules known as opsins, which also exist in the retina and are known to be photosensitive. They are hypothesized to play a role in distributed light sensing and control in the periphery~\cite{mathger2010evidence}, thus potentially adding an additional distributed component for skin patterning that enables sensing and actuation independent of the brain. In~\cite{kingston2015visual}, the authors proposed three hypotheses for this kind of distributed light-sensing system. First, sensing by chromatophores could trigger its expansion and retraction by changing a single component like radial muscles. Second, a patch of chromatophores could respond to light stimuli as a unit, as local receptors could communicate with each other among the chromatophores~\cite{cloney1968ultrastructure}. Finally, signals produced by chromatophores may travel by afferent nerve fibers to the central nervous system to provide additional information about the environment in which the animal exists.

\section{Mathematical Models for Pattern Formation}
\label{sec:model}

We have discussed different ideas of how pigment cell patterns are formed on skins, how the pigment cells are controlled locally and globally on the cellular or molecular level, and how they communicate with each other. In this section, we review the possible mathematical models that can describe pattern formation.

Albeit morphological and physiological pattern formation are fundamentally distinct processes, the resulting patterns are remarkably similar (see Figure \ref{fig:camouflage_patterns}). The dominating shapes are stripes and mottles or combinations thereof, suggesting common mathematical models to describe them, which we will discuss further below. Indeed, the majority of mathematical models for pattern formation are concerned with (semi-)static patterns as observed in morphological pattern formation, not limited to vertebrates, but also observed on sea shells, sand dunes, and plants. 

In this paper, we review the reaction-diffusion model and variants thereof as well as a cellular automata, a general version of discretized reaction-diffusion models.

\subsection{Reaction-diffusion Model}

One of the earliest models for pattern formation is the reaction-diffusion model that involves short-range positive feedback and long-term negative feedback~\cite{turing1952chemical, meinhardt2009algorithmic, kondo2010reaction}. In 1952, Alan Turing published his now classic paper describing the chemical process between signaling molecules that spread away from their source to form a concentration gradient (``morphogens'') within a series of cells~\cite{turing1952chemical}. Turing’s basic idea is that ``the mutual interaction of elements results in spontaneous pattern formation''. He suggested that morphogens could react with each other and diffuse through cells forming patterns through the reaction-diffusion process.%, which is the so-called ``reaction-diffusion model''. 
This combination of positive and negative feedback results in a large variety of patterns. The resulting stationary patterns are called Turing pattern, which is a kind of nonlinear wave that is maintained by the dynamic equilibrium of the system~\cite{kondo2010reaction}. Most of the patterns seen in nature can be replicated by this kind of model.
%Indeed, uniform, mottled, and disruptive patterns are ubiquitous in nature, not limited to camouflage patterns, but can also be observed in crabs or seashells. Those patterns and can all be described with very similar mathematical tools~\cite{meinhardt2009algorithmic}. 
In biology, one of the elementary processes in morphogenesis is the regulated formation of a spatial pattern of tissue structures, starting from almost homogeneous tissue~\cite{gierer1972theory}. 
%The best-known theoretical model used to explain self-regulated pattern formation is the reaction-diffusion model or Turing model.  Due to instabilities in the system, the morphogens both react and diffuse, which changes their concentration within each cell.

%The resulting Turing patterns from various reaction specifications have been shown to replicate most of the biological spatial patterns, including regular patterns reminiscent of those found on many animals~\cite{kondo2010reaction}. 
Furthermore, Turing patterns have the ability to self-regulate the patterns and exhibit robustness against perturbation~\cite{barrass2006mode}. This ability helps to explain the autonomy shown by pattern-forming developmental processes~\cite{meinhardt2009algorithmic}. For example, the size of spots or width of stripes will not change, but new spots or stripes will appear to extend the space if in simulation the field is enlarged by adding more cells in which reaction occurs~\cite{yamaguchi2007pattern}. Hereafter, Gierer and Meinhardt proposed that Turing patterns can be formed by only involving two different feedback mechanisms, a short-range positive feedback and a long-range negative feedback~\cite{gierer1972theory,meinhardt2000pattern}. The positive and negative feedback loops are also called activator and inhibitor respectively. This is now accepted as the basic requirement for Turing pattern formation~\cite{koch1994biological, gierer1972theory, meinhardt1982models,meinhardt2000pattern,meinhardt2009algorithmic}. %Read~\cite{kondo2010reaction} for more details about reaction-diffusion models for biological pattern formation.

The hypothetical molecules in the original reaction-diffusion model are idealized for the purposes of mathematical analysis. The hypothesis is that the elementary process in pattern formation may be the formation of a primary pattern of two morphogens, one acting as activator, and one with inhibitory effect, the inhibition being derived from, and extending into a wider area~\cite{gierer1972theory}. We now describe the basic of reaction-diffusion models and some of their characteristics, where two morphogens interact via a set of nonlinear partial differential equations:

\begin{equation}
\frac{\partial u}{\partial t} = F(u, v) + D_u \Delta u
\end{equation}
\begin{equation}
\frac{\partial v}{\partial t} = G(u, v) + D_v \Delta v
\end{equation}

where $u$ and $v$ are the morphogen concentrations; $F(u, v)$ and $G(u, v)$ are the functions controlling the production rate of $u$ and $v$ (see~\cite{sanderson2006advanced, meinhardt2009algorithmic} with various forms of the two functions for different application of interest); $D_u$ and $D_v$ are the diffusion rates, and  $\Delta u$ and  $\Delta v$ are the Laplacians of $u$ and $v$ representing isotropic diffusion.

In practice, all simulations, in Figure~\ref{fig:model_realsystem}, of reaction-diffusion models are performed by solving the set of equations through discretizing on both time and space. The discrete reaction-diffusion model can also be seen as a cellular automaton~\cite{wolfram1983statistical, wolfram1984cellular}. Cellular automata are grids of cells whose states are iterately updated based on the states of neighboring cells according to a set of rules. 

%In the next section, We will show simulation results of reaction-diffusion models and cellular automata and how they compare with patterns exhibited by animal models.

\subsection{Reaction-diffusion Models on Real Systems}
\label{sec:realsystem}

\begin{figure*}[!htb]
\centering
\includegraphics[width=.92\textwidth]{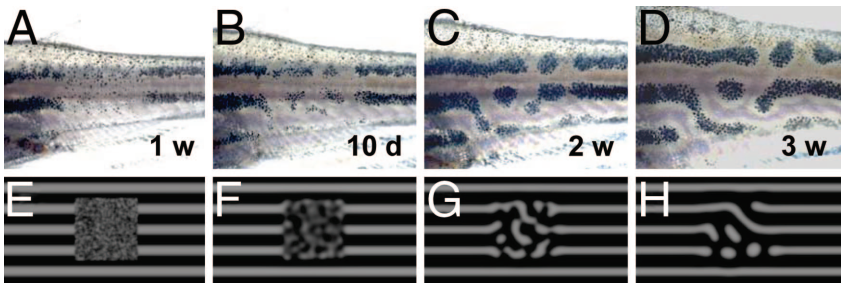}
\\
\includegraphics[width=.3\textwidth]{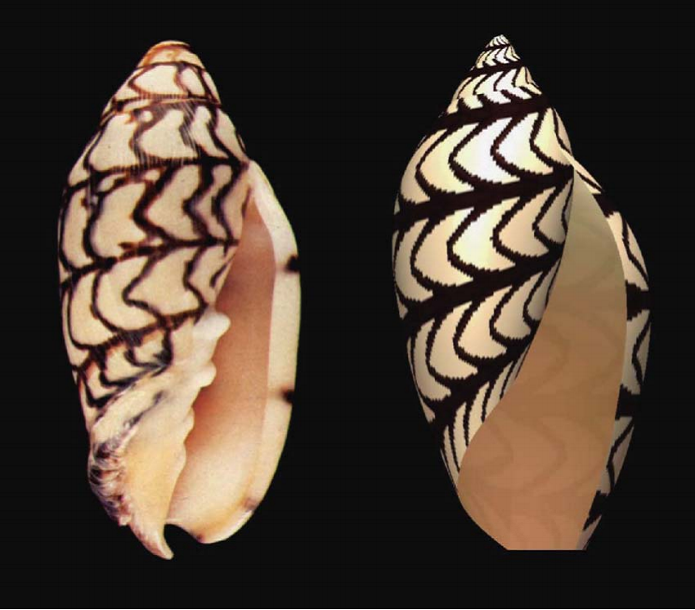}
\includegraphics[width=.3\textwidth]{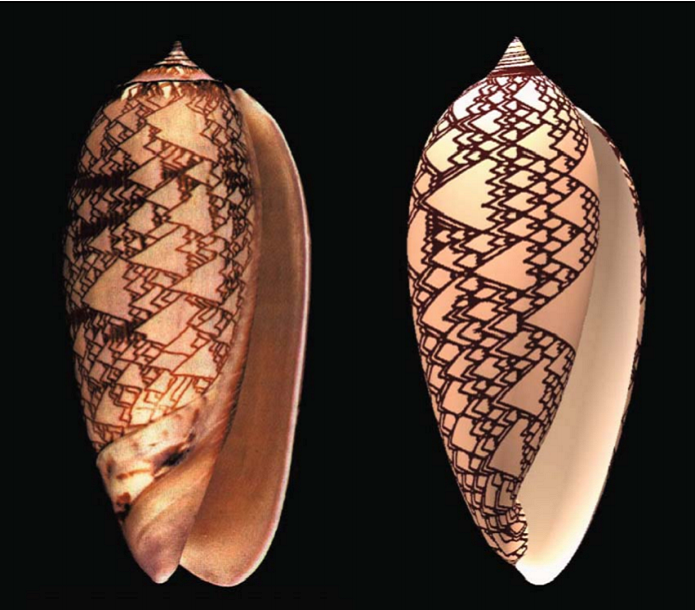}
\includegraphics[width=.3\textwidth]{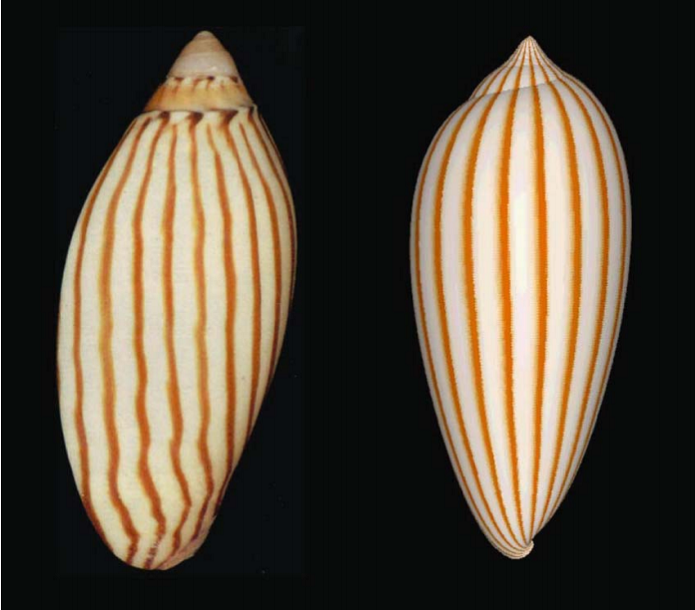}
\\
\includegraphics[width=.3\textwidth]{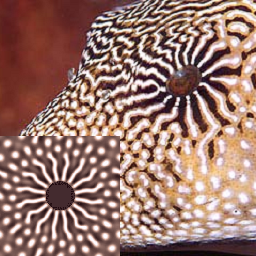}
\includegraphics[width=.3\textwidth]{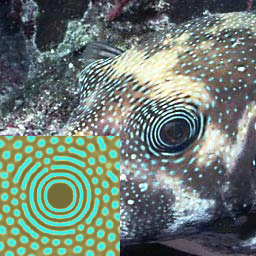}
\includegraphics[width=.3\textwidth]{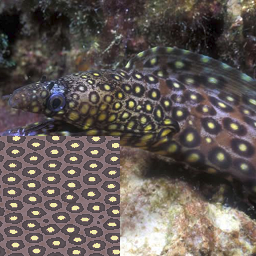}
\caption{(A) Regeneration process of the stripes of both zebrafish and computer simulation~\cite{yamaguchi2007pattern}.
(B) Synthetic images of selected shells. Left: model of Volutoconus bednalli (Bednall’s Volute). Middle: model of Oliva porphyria. Right: model of Amoria ellioti~\cite{meinhardt2009algorithmic}.
(C) Examples of kinetics/systems to match patterns found in nature. Top row: circular stripe-spot pattern on blue spotted puffer fish. Middle row: radial stripe-spot pattern on map toby puffer fish. Bottom row: double spot pattern on jewel moray eel~\cite{sanderson2006advanced}.}
\label{fig:model_realsystem}
\end{figure*}

%Over the past three decades, most pattern formation models are proposed based on the ``short-range activation and long-range inhibition'' interaction between two different concentrations and are applied in different application areas, like mathematical biology~\cite{kondo2010reaction}, chemical dynamics~\cite{epstein1998introduction} and computer graphics~\cite{sanderson2006advanced}.

We will now describe animal studies that validate the proposed generative models. In addition to provide qualitatively similar results, additional evidence is obtained by introducing defects and compare the repair process in the mathematical and the animal model. 

Fish pigment pattern development is an ideal system to study the reaction–diffusion mechanism because the pattern formation occurs in the two-dimensional field, and the dynamics of this pattern are trackable. Zebrafish is a commonly used model system. The common way in research is to ablate pigment cells in certain part of fish skins and observe the development of new pigment cells and regeneration of patterns. In~\cite{yamaguchi2007pattern}, the authors researched the pattern regulation and regeneration in the stripe of zebrafish by ablating the pigment cells in limited areas of zebrafish skins, and the experiments revealed that the mechanism underlying the pattern formation of zebrafish is highly dynamic and autonomous. The regenerated patterns and the transition of the stripes during the regeneration process suggest that pattern formation is independent of the pre-pattern, which is also a characteristic property of an reaction-diffusion model. The comparison between the regenerated patterns and the patterns from computer simulations strongly suggested that the stripe patterning is based on an autonomous mechanism such as an reaction-diffusion system by (see Figure~\ref{fig:model_realsystem}, top row). Furthermore, in~\cite{nakamasu2009interactions}, the authors employed similar methods to explore the mechanism in cellular level, and they found that interactions between zebrafish pigment cells responsible for the generation of Turing patterns. %The stripes of zebrafish are composed of three types of pigment cells: melanophores (black), xanthophores (yellow), and iridophores (white). 
They observed the in-vivo interactions between black (melanophores) and yellow (xanthophores) cells with respect to their distance and how these interactions affect the development and survival of other pigment cells. This observation suggests that the development of melanophores was positively affected by xanthophores in the neighboring stripes. The authors in~\cite{nakamasu2009interactions} also found that the development and survival of the cells were influenced by the positioning of the surrounding cells. However, the mechanism underlying pattern formation remains unknown because the molecular or cellular basis of the phenomenon has yet to be identified. To test reaction-diffusion models, the authors in paper~\cite{muller2012differential}  measured the biophysical properties of the Nodal/Lefty activator/inhibitor system during zebrafish embryogenesis. These results indicate that differential diffusivity is the major determinant of the differences in Nodal/Lefty range and provide biophysical support for reaction-diffusion models of activator/inhibitor-mediated patterning.

There is also research on designing reaction-diffusion models to generate patterns seen on animals like seashells and fish. The patterns obtained from simulation look very similar to those compared patterns on real lives. Patterns in seashells result from the deposition of pigmented material at the shell margin. In~\cite{meinhardt2009algorithmic}, the authors illustrate this process using synthetic images of selected seashells (see Figure~\ref{fig:model_realsystem}, middle row). They created a comprehensive model of seashells that would incorporate patterns into three-dimensional shell shapes. They solved differential equations representing pigment deposition along the edge of the seashell. The pattern unfolds on the seashell surface as the shell grows. In ~\cite{page2005complex}, the authors explored the effects of spatially varying parameters on pattern formation in one and two dimensions using the Gierer–Meinhardt reaction-diffusion model. In ~\cite{sanderson2006advanced}, the authors explored different reaction-diffusion models for texture synthetics using computer graphics (see Figure~\ref{fig:model_realsystem}, bottom row). They tried to control the system by analyzing related parts like system instabilities, parameter mapping, reaction kinetics, and diffusion kinetics. To form the desired pattern, they start from one basic model and keep adding other models until the expected pattern forms. Unless we understand the characteristics of each component, it would be difficult to generate a specific pattern. In addition, to form a complicated pattern, they explored using two models simultaneously and allowing them to diffuse.

\begin{figure}[!htb]
\centering
\includegraphics[width=.91\columnwidth]{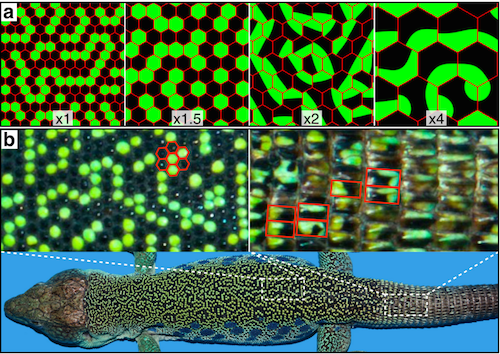}
\\
\includegraphics[width=.3\textwidth]{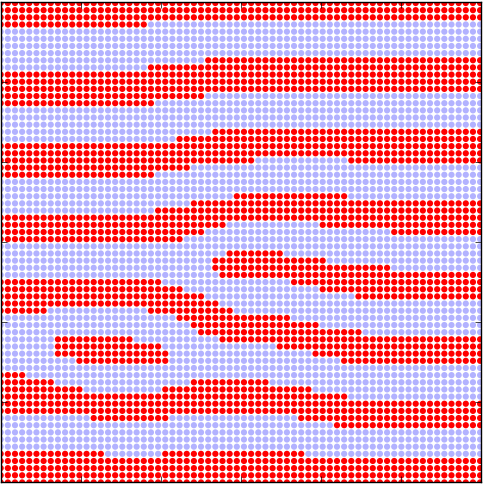}
\includegraphics[width=.3\textwidth]{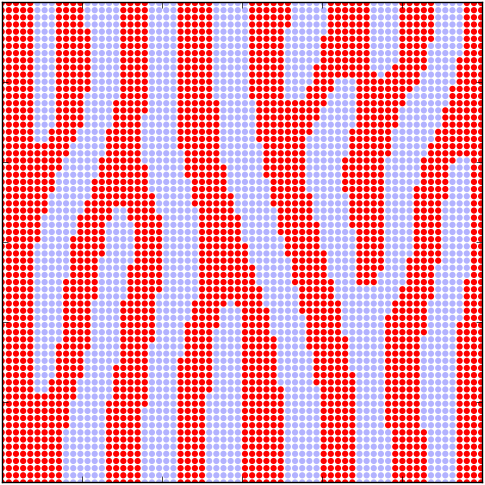}
\includegraphics[width=.3\textwidth]{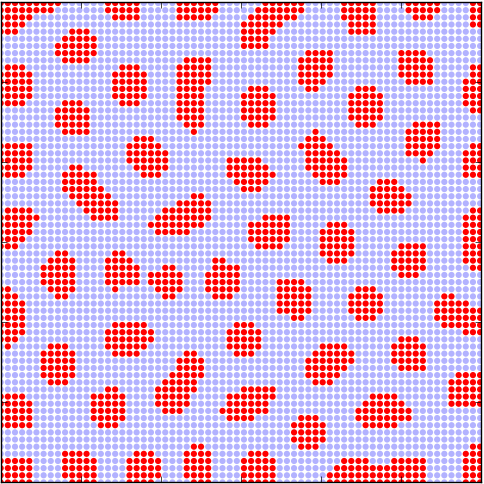}
\caption{Patterns formed with cellular automata.
Top row: (a) patterns generated by reaction-diffuion based cellular automata with different scale sizes and (b) ocellated lizard skin colour patterns  \cite{manukyan2017living}.
Bottom row: stripe and spot patterns generated by a local activator-inhibitor model~\cite{young1984local}.}
\label{fig:cellularautomata}
\end{figure}

Recently, it was shown that cellular automata are also directly applicable to natural systems. In paper~\cite{manukyan2017living}, the authors tracked time series of ocellated lizard scale color dynamics over four years of lizards' development. They found that skin patterns could be produced by a cellular automaton that dynamically computes the color states of individual pigment cells to produce the color pattern (see Figure~\ref{fig:cellularautomata}). Using numerical simulations and mathematical derivation, they identify how a discrete cellular automaton emerges from a continuous reaction–diffusion system. Their study indicates that cellular automata can directly correspond to processes generated by biological evolution. 

Another discrete, local activator-inhibitor model for pattern formation that assumes only local cell interactions is known as Young's model cite{young1984local}. In the model, activator and inhibitor mophorgens are defined with different shape around each pigment cell. With specifically definded morphogens, Young's model can generate mottles of different sizes or stripes of different thickness and directions (Figure \ref{fig:cellularautomata}, bottom row).

As cellular automata are fully distributed --- that is, each state is only influenced by the state of its neigbhors, this property allows us to infer the required communication in a physical system for either understanding biological systems or engineering artificial ones.

%%%%%%%%%%%%%%%%%%%%%%%%%%%%%%%%%%%%%%%%%%%%%%%%
% Camouflage Materials inspired by Animals
%%%%%%%%%%%%%%%%%%%%%%%%%%%%%%%%%%%%%%%%%%%%%%%%

\section{Camouflage Materials Inspired by Animals}
\label{sec:material}

After reviewing the underlying mechanisms and algorithms of natural camouflage systems, we will now review the state of the art in engineering camouflage systems.
There have been multiple attempts to achieve adaptive camouflage using a combination of cameras and projection~\cite{inami2003optical,lin2009framework,lasbury2017cloaking}. In these works, the background image is usually caught by a camera and projected on the front surface of an object. The surface is made of retro-reflective material to reflect the most light projected on it. In this way, an object can become virtually invisible from the viewpoint of an observer. Although such systems provide ``perfect'' camouflage, they are highly dependent on the observer's viewpoint. They are therefore impractical in scenarios where the position of the observer is unknown or where there are multiple observers. A concept often confused with camouflage is that of ``cloaking'' using optical metamaterials, which allow manipulating light waves in a way that allows them to be routed around an obstacle\cite{pendry2006controlling}. Albeit partial invisibility has been achieved even for larger objects~\cite{smith2014cloaking}, existing approaches are usually limited to light of a very specific wavelength. Both approaches are distinctly different from active camouflage in animals, which neither simply reproduce an observed pattern nor employ cloaking techniques.

%These approaches are radically different from the mechanisms of animal camouflage discussed above. 
There exist a few attempts to create artificial chromatophores, camouflage skins, or entire adaptive camouflage systems that are inspired by animal models. We shall find that the large majority of works focuses only on sub-problems of the camouflage systems, like, individual devices, mechanisms, or algorithms that often ignore the system-level challenges of the camouflage problem.

\subsection{Artificial Chromatophores}

\begin{figure}[!htb]
\centering
\includegraphics[width=.22\textwidth]{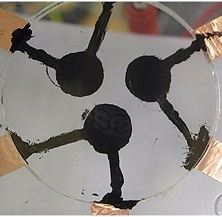}
\includegraphics[width=.22\textwidth]{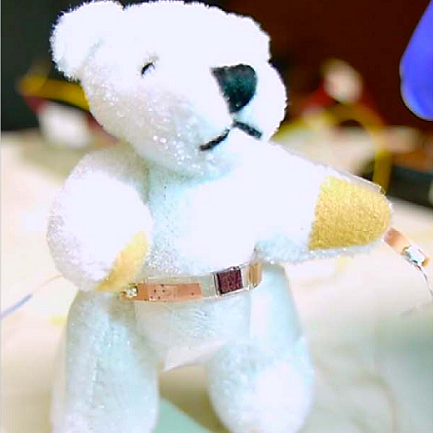}
\includegraphics[width=.22\textwidth]{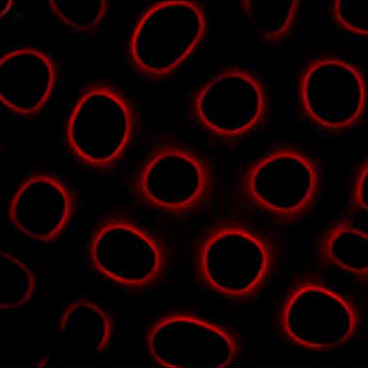}
\includegraphics[width=.22\textwidth]{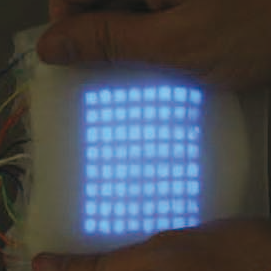}
\\
\includegraphics[width=.22\textwidth]{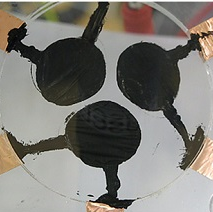}
\includegraphics[width=.22\textwidth]{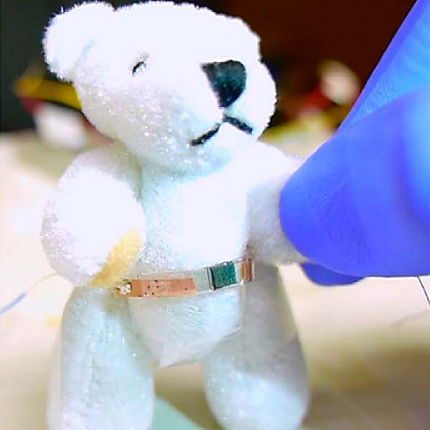}
\includegraphics[width=.22\textwidth]{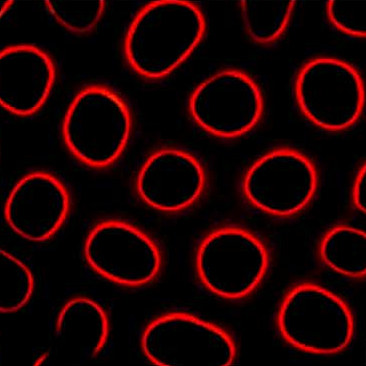}
\includegraphics[width=.22\textwidth]{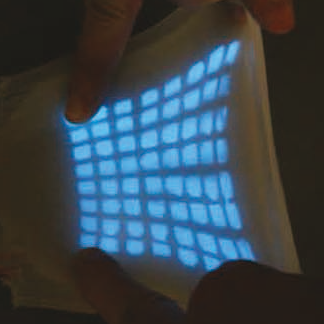}
\caption{A list of artificial chromatophores.
Column 1: Biomimetic chromatophores: three prototype artificial chromatophores are shown in unactuated (left) and actuated (right) states~\cite{fishman2015hiding}.
Column 2: An interactive color-changing and tactile-sensing e-skin. Sequential images of a teddy bear show the expression of tactile sensing into visible color changes~\cite{chou2015chameleon}.
Column 3: On-demand fluorescent patterning, where the applied voltages induce various patterns of large deformation on the surface, which subsequently exhibits corresponding fluorescent patterns~\cite{wang2014cephalopod}.
Column 4: Multipixel electroluminescent displays fabricated via replica molding~\cite{larson2016highly}.}
\label{fig:material_chromatophore}
\end{figure}

% These materials belong to the new class of high-performance materials, commonly known as smart materials, developed by chemists, physicists, materials engineers.

There are various categories among color-changing or chromogenic materials, such as photochromic, thermochromic, and electrochromic materials, which take their names from the energy source that provokes the modification of optical properties. Among them, electrochromism is the most versatile of all chromogenic technologies because it is the easiest to control and it can be used in combination with different stimuli such as stress or temperature \cite{marinella2013materials}. Here in this section, we report several artificial chromatophores that change color provoked by electricity.

In~\cite{rossiter2012biomimetic}, the authors present a soft and compliant artificial chromatophore that mimics the contrasting mechanisms of sacculus expansion employed by cephalopod chromatophores using electroactive polymer artificial muscles (see column 1 in Figure~\ref{fig:material_chromatophore}). Also inspired by cephalopods, % that can display dazzling patterns,
in paper~\cite{wang2014cephalopod}, the authors designed an electro-mechano-chemically responsive elastomer skin (see column 2 in Figure~\ref{fig:material_chromatophore}). Voltage can cause the deformation on the surface of the skin, and therefore lead to different fluorescent patterns including lines, circles, and letters on demand. However, unlike the skin presented in~\cite{rossiter2012biomimetic} where the cells can be controlled independently, the whole skin is controlled by one central source so that the skin can only show one pre-defined pattern.

Inspired by the chameleon, the authors in~\cite{chou2015chameleon} propose a stretchable e-skin that can change color on demand of pressure (see column 3 in Figure~\ref{fig:material_chromatophore}). Each part of e-skin composes two parts, a resistive pressure sensor and the organic electro-chromic devices (ECD). The resistive pressure sensor is used to mimic the pressure-sensing properties of natural skin. The ECD has the advantage of color retention being called the `color memory effect'. By simply applying various pressure on the pressure sensor, the color of the ECD can be controlled. Since the e-skin is not designed for camouflage, it can't be utilized to integrate with pattern formation algorithm directly. But we can easily control the color change of each part of the skin by applying different voltage with micro-controllers.

In ~\cite{larson2016highly}, the authors present a stretchable electroluminescent skin for display(see column 4 in Figure~\ref{fig:material_chromatophore}). The skin is composed of an array of pixels, each of which can also be independently controlled `on' or `off'. The dynamic control of the skin is demonstrated to work for displaying different simple patterns. In addition to emitting light, pixels can act like distributed sensors as they can sense deformations from pressure and stretching, which will provide extra environmental information to the system.

\subsection{Artificial Camouflage Systems}

\begin{figure}[!htb]
\centering
\includegraphics[width=.3\textwidth]{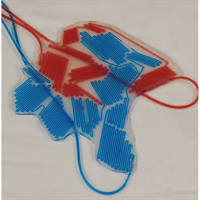}
\includegraphics[width=.3\textwidth]{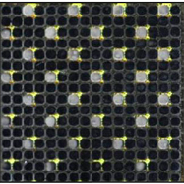}
\includegraphics[width=.3\textwidth]{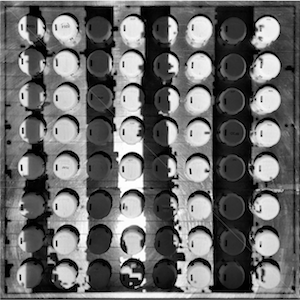}
\\
\includegraphics[width=.3\textwidth]{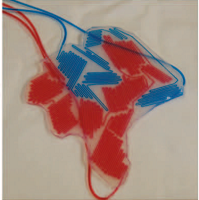}
\includegraphics[width=.3\textwidth]{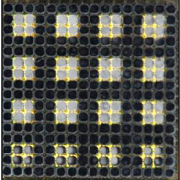}
\includegraphics[width=.3\textwidth]{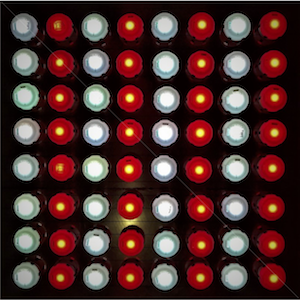}
\caption{A list of camouflage skins or systems.
Column 1: Camouflage and display for soft machines~\cite{morin2012camouflage}.
Column 2: Adaptive camouflage systems that incorporate essential design features found in the skins of cephalopods: illustration of metachrosis for several different static patterns~\cite{yu2014adaptive}.
Column 3: Distributed camouflage system for smart materials~\cite{li2017distributed}.}
\label{fig:material_system}
\end{figure}

Different systems are designed for adaptive camouflage in the literature, and we review some of them that employ a variaty kind of pattern formation schemes.

Authors of~\cite{rossiter2012biomimetic} continued the work of artificial chromatophores (see column 1 in Figure~\ref{fig:material_chromatophore}) by proposing an artificial chromatophore system that mimics the functionality of cephalopods' skin, especially the pattern generation technique~\cite{fishman2015hiding}. The artificial skin is made from electroactive dielectric elastomer: a soft, planar-actuating material that mimics the actuation of biological chromatophores. The dynamic pattern generation can be achieved by imposing simple local rules into the artificial chromatophore cells where each cell changes its status based on its neighbors. The skin could be programmed with specific parameters of the proposed model to generate a variety of controllable patterns such as unidirectional propagation and sawtooth-like oscillation. However, the authors didn't program the proposed skin with their model that is proved working in simulations.

A multi-layered flexible sheet, which tries to mimic the three-layered functionality of the cephalopods' skin, is presented in paper~\cite{yu2014adaptive} (see Figure~\ref{fig:material_system}E).  The multilayer stack includes a color-changing element (analogous to a chromatophore), an actuator (analogous to the muscles that control the chromatophore), and a light sensor (analogous to a functional unit involving opsins). The working mechanism is that the actuator is activated by heat, leading to the color-changing element changing from black to white, when there is light on the light sensor. For details of those components, see the original paper. The work combines distributed sensing and actuation with coloration to provide the adaptive camouflage functionality of the cephalopod skin.

The authors of \cite{morin2012camouflage} designed a material embedded with microfluidic networks for soft machines to camouflage or display (see column 2 in Figure~\ref{fig:material_system}). Those microfluidic networks are contained in thin silicon sheets. Camouflage or display of the material are controlled by pumping colored or temperature-controlled fluid through the networks. %Why is this a ``system'' and not an ``artificial chromatophore''}
The authors attached the sheet to the top of soft machines to also demonstrate the lightness and flexibility. It is shown by experiments that the presented material can help soft machines to camouflage in different environments like rock bed or leaf-covered slab, although not perfectly. However, they also point out that advanced autonomous systems are demanded for some applications. For example, one component is needed for determining color and pattern of surrounding environment so that the material can adaptively change the display. For the hardware design, although the material can change its color by pumping into different colored fluid, its pattern is fixed. In other words, this material cannot self-organize patterns. We can manually replace the ``skins'' of a soft machine to change patterns, however, this effectively limits the application range of the material.

The authors of \cite{li2017distributed} presented a distributed camouflage system (see column 3 in Figure~\ref{fig:material_system}) with a swarm of 64 static miniature robots (“Droplets”) to camouflage in an environment by generating colored patterns similar to those perceived in the environment. Each particle they used is equipped with sensing, computation, and local communication abilities. The system consists pattern recoginition, pattern formation, and consensus algorithm. Together, these algorithms enable the swarm to obtain a high-level understanding of its environment and to quickly adapt its appearance to changing environments.

However, very few works articulate the systems challenges that require not only local color changes but also local sensing and computation or investigate the ability to co-locate simple signal processing with the sensors themselves~\cite{fekete2009distributed}.

\section{Discussion}
\label{sec:discussion}

How to design an dynamic camouflage system is the main topic of this survey. We have listed some recent research that tries to mimic the dynamic camouflage capability of cephalopods or chameleons in Section~\ref{sec:material}. There is work on understanding the mathematics of pattern formation, designing artificial chromatophores, some of which integrated into skins that can display different patterns. However, the existing body of work exclusively focuses on components of camouflage systems. Albeit one might argue that what remains is a system integration problem, we argue that the majority of components reviewed have not been designed with system integration mind. For example, applying a distributed pattern formation algorithm on an artificial skin requires the cells in the skin to have the ability to communicate with neighboring cells to avoid a central controller for global information collection, reasoning, and distribution. Although our own work \cite{li2016distributed} comes closest in that it presents a fully distributed cellular system for pattern recognition and formation, it fails to address passive coloration change as well as the problem of powering such a system. Similarly, camouflage systems with passive color change such as \cite{morin2012camouflage} ignore challenges of sensing, computation, and distributed actuation, i.e., the ability to locally change coloration. Albeit closest to the functionality of a biological chromatophore, we believe that physically revealing pigment by either local actuators or pumps will remain a hard mechanical challenge that will likely be substituted by electrochromic materials that can be activated by light, temperature, electric field, or a combination thereof. 

Where computation to process the environment and generate appropriate patterns happens is a question that remains equally vexing to the biology and engineering communities. From an engineering perspective, this is a trade-off between computation and communication. Calculating the pixel values for a specific camouflage pattern on a skin in a central location (akin to the brain) requires transporting this information throughout the body. This is in contrast to generating patterns locally, which requires local computation, but less information to communicate. A compelling argument for pattern generation to be local can be made by looking at the information entropy \cite{shannon2001mathematical} of Turing patterns. Albeit very high at first sight, all mathematical models reviewed here require only very few parameters to fully define a pattern, including linear combinations thereof. Together with the fact that generating these patterns locally requires only local information, motivates a local generation of such patterns. Albeit we cannot speak for natural systems, addressing millions of pixels on an amorphous 3D surface and routing information there from a central location will remain a challenge from an engineering perspective \cite{correll2017new}.

On the other hand, performing local reasoning using a distributed system is not very efficient beyond recognizing very simple patterns \cite{otte2016collective}. Indeed, it is commonly believed that animals see and think using their brain, and are able to perform high-level, semantic scene understanding. This is now possible with state-of-the-art convolutional neural networks (CNN), which can correctly classify images with a high chance. CNNs are a class of algorithms where connectivity pattern between its neurons are inspired by the organization of the animal visual cortex. Similarly, it is now possible to generate arbitrary images using transposing CNN, that use a signal vector --- such as a high-level scene description --- as the input and a pattern image as the output, instead of an image as the input and a vector as the output in CNN. In \cite{radford2015unsupervised}, a transposed CNN to generate images is trained using a second CNN that is used as a discriminator to evaluate the generated images, in turn providing information to improve the generator. Such an approach would map naturally to a biological context, for example by training generating functions that make detection in a certain environment hardest. 

Recent advances in machine learning might also help with addressing trade-offs in computation and communication by integrating the communication structure into the learning problem. For example, \cite{hughes2016distributed} proposes to integrate computational synapsis with bandwidth and time constraints into a CNN framework to find appropriate trade-offs between computation and communication given specific available communication channels. 

%There are some challenges for centralized pattern formation. The first one is building a signal network. We need to simulate the network, through which nervous signals transfer from brain to every cell in the skin. The second challenge is physical implementation. For example, which connection method should be used, wired or wireless. Wired connection is straightforward but the connection is cumbersome, whereas wireless connection has problems with identification, frequency, and workload. The last challenge is that both wired and wireless connections need to localize each cell since patterns are formed by the color change of corresponding positional pigment cells. There are two challenges for distributed pattern coloration. The first one is to discretize a certain reaction-diffusion model (reaction-diffusion model is currently the most proven model for biological pattern formation) to make it work in a distributed way. After the discretized model is prepared, the artificial skin should be designed accordingly for integration with the model, which means the skin as a whole can receive the command from the AI system and can generate an expected pattern with the discretized model in a distributed way.

Other aspects in deciding whether to perform computation centrally or distributed are scalability and robustness. In a centralized framework, a central computing entity needs to scale up with the number of artificial chromatophores. (Indeed, brain-mass in mammals is proportional to their size due to the challenge of controlling a larger body.) In a decentralized framework, computing power is bound by the requirements of selecting an appropriate pattern. A centralized framework is robust to the failure of individual chromatophores, but fails catastrophically when the centralized computing element fails. A decentralized framework would be robust only when perception is also performed in a decentralized way. Here, we recall that it is not fully established yet in how far cephalopods rely on distributed sensing.

%Choosing either centralized or distributed method influence how we design and control each piece of the color changing material, in particular if each piece has the ability to compute independently or to communicate locally or globally. In a centralized scheme, it is an overloaded work for a center to keep the connection with all the cells, especially as the number increases, while each cell in a decentralized scheme needs only to connect with a few neighbor cells. The advantages for the centralized scheme are the simple central control algorithms and the direct formation of expected patterns with stable connections, while the advantages of the distributed scheme are the scalablity to the number of cells and the robustness to failures of connections or cells.

\section{Conclusion}
\label{sec:conclusion}
Biological camouflage systems are providing animals with amazing capabilities that motivate us to imitate and emulate them. For now, the research of artificial camouflage systems is still at its early stage.
We systematically researched the components for designing an artificial camouflage system, including sensing, computation, communication, and actuation, from a system-level perspective. %Many methods are used in nature to accomplish camouflage, such as changing body shape, changing skin texture, and changing skin pattern. Here we selected camouflage with pattern formation as our main focus. 
We differentiate static and adaptive camouflage and explain how they are implemented in natural systems. We  describe possible mathematical models for pattern formation --- the reaction-diffusion model, which is proposed mainly for the pattern formation happened during growth, assuming short-range positive feedback and long-range negative feedback, and the interactions between neighboring cells. %We listed research that proves the reaction-diffusion models can replicate most of the biological skin patterns. 
We also describe component and systems to engineer artificial camouflage systems. We observe that the existing body of work exclusively focuses on components, and does not address the system challenges of camouflage.

%Finally, we discussed the problems to be solved for implementing an artificial adaptive camouflage system and proposed one possible way of making such a system.

Implementing an artificial camouflage system remains a hard challenge since the biological mechanisms underlying animals' camouflage are not fully understood yet. Also, the problem requires tight collaboration between disparate areas such as mathematics, material science, computer science, and electrical engineering, among others. The involved trade-offs, in particular where in the body sensing and computation should take place, are similar in both engineering and natural evolution. It might therefore be likely that advances in engineering will also generate plausible hypothesis for animal physiology.  We hope that this survey provides other researchers a systematic perspective of designing such a system and stimulate such inter-disciplinary research.

%\printbibliography[heading=bibintoc]
\printbibliography %added

\end{document}